\documentclass[sigconf]{acmart}
\usepackage{algorithm}
\usepackage{algorithmic}
\usepackage{listings}
\usepackage{CJKutf8}
\usepackage{fancyvrb}
\usepackage{latexsym}
\usepackage{xspace}
\usepackage{booktabs}
\usepackage{graphicx}
\usepackage{bbding}
\usepackage{multirow}
\usepackage{balance}
\usepackage{longtable}

\usepackage[utf8]{inputenc}
 
\usepackage{amssymb}
\usepackage{amsmath}
\usepackage{bm}
\usepackage{microtype}
\usepackage{array}
\usepackage{listings}
\usepackage{tcolorbox}
\usepackage{geometry}
\usepackage{longtable} % 支持跨页的长表格
\newcommand{\ourmodel}{{SynthRAG}\xspace}

\AtBeginDocument{%
  }

\renewcommand\footnotetextcopyrightpermission[1]{}
\setcopyright{none}

\setcopyright{acmlicensed}
\copyrightyear{2018}
\acmYear{2018}
\acmDOI{XXXXXXX.XXXXXXX}

\acmConference[Conference acronym 'XX]{Make sure to enter the correct
  conference title from your rights confirmation emai}{June 03--05,
  2018}{Woodstock, NY}
\acmISBN{978-1-4503-XXXX-X/18/06}

\begin{document}

% \title[]{A Systematic Retrieval-Augmented Framework for Explanatory Answer Generation}

\title[]{An Adaptive Framework for Generating Systematic Explanatory Answer in Online Q\&A Platforms}

\author{Ziyang Chen}
\orcid{0000-0002-1714-0304}
\affiliation{%
 \institution{Independent Researcher}
 \city{Changsha}
 \country{China}
}
\email{chenziyang319@163.com}

\author{Xiaobin Wang}
\affiliation{%
 \institution{Alibaba Group}
 \city{Hangzhou}
 \country{China}
}
\email{xuanjie.wxb@alibaba-inc.com}

\author{Yong Jiang}
\affiliation{%
 \institution{Alibaba Group}
 \city{Hangzhou}
 \country{China}
}
\email{yongjiang.jy@alibaba-inc.com}

\author{Jinzhi Liao}
\affiliation{%
 \institution{Independent Researcher}
 \city{Changsha}
 \country{China}
}
\email{jinzhiliao19@163.com}

\author{Pengjun Xie}
\affiliation{%
 \institution{Alibaba Group}
 \city{Hangzhou}
 \country{China}
}
\email{chengchen.xpj@alibaba-inc.com}

\author{Fei Huang}
\affiliation{%
 \institution{Alibaba Group}
 \city{Hangzhou}
 \country{China}
}
\email{feirhuang@gmail.com}

\author{Xiang Zhao}
\affiliation{%
 \institution{Independent Researcher}
 \city{Changsha}
 \country{China}
}
\email{xiangz@aliyun.com}
 
% \renewcommand{\shortauthors}{Anonymous Author, et al.}


\begin{abstract}
%In the era of information explosion, 
Question Answering (QA) systems face challenges in handling complex questions that require multi-domain knowledge synthesis. 
The naive RAG models, although effective in information retrieval, struggle with complex questions that require comprehensive and in-depth answers.
The pioneering task is defined as explanatory answer generation, which entails handling identified challenges such as the requirement for comprehensive information and logical coherence within the generated context.
To address these issues, we refer to systematic thinking theory and propose SynthRAG, an innovative framework designed to enhance QA performance. 
SynthRAG improves on conventional models by employing adaptive outlines for dynamic content structuring, generating systematic information to ensure detailed coverage, and producing customized answers tailored to specific user inquiries. 
This structured approach guarantees logical coherence and thorough integration of information, yielding responses that are both insightful and methodically organized.
Empirical evaluations underscore SynthRAG's effectiveness, demonstrating its superiority in handling complex questions, overcoming the limitations of naive RAG models, and significantly improving answer quality and depth. Furthermore, an online deployment on the Zhihu platform revealed that SynthRAG's answers achieved notable user engagement, with each response averaging 5.73 upvotes and surpassing the performance of 79.8\% of human contributors, highlighting the practical relevance and impact of the proposed framework. Our code is available at \url{https://github.com/czy1999/SynthRAG}.

\end{abstract}

\begin{CCSXML}
<ccs2012>
   %     <concept_id>10002951.10003317.10003325</concept_id>
   %     <concept_desc>Information systems~Information retrieval query processing</concept_desc>
   %     <concept_significance>500</concept_significance>
   %     </concept>
   <concept>
       <concept_id>10002951.10003227</concept_id>
       <concept_desc>Information systems~Information systems applications</concept_desc>
       <concept_significance>500</concept_significance>
       </concept>
 </ccs2012>
\end{CCSXML}

% \ccsdesc[500]{Information systems~Information retrieval query processing}
\ccsdesc[500]{Information systems~Information systems applications}

\keywords{Retrieval-Augmented Generation, Explanatory Answer Generation, Information Integration, Large Language Models}


\maketitle

\section{Introduction}

In the era of information explosion, Question Answering (QA) systems have become an indispensable component within the realm of
artificial intelligence~\cite{chen-etal-2023-multi,Saxena2021QuestionAO,DBLP:journals/corr/abs-2311-09149}, widely employed in search engines, intelligent assistants, and domain QA platforms, greatly improving the efficiency and accuracy of information retrieval~\cite{DBLP:journals/corr/abs-2404-14851}. 
However, with the increasing demands of users, practical questions in daily life are progressively more complex. 
In particular, users require an explanatory context for a given question rather than an entity, phrase, or sentence.
To meet the requirement, it is imperative to integrate multiple sources of information to provide insightful answers~\cite{DBLP:conf/sigir/LuPRAWW19,li2024uni}.

\begin{figure}[t]
  \centering
  \includegraphics[width=\linewidth]{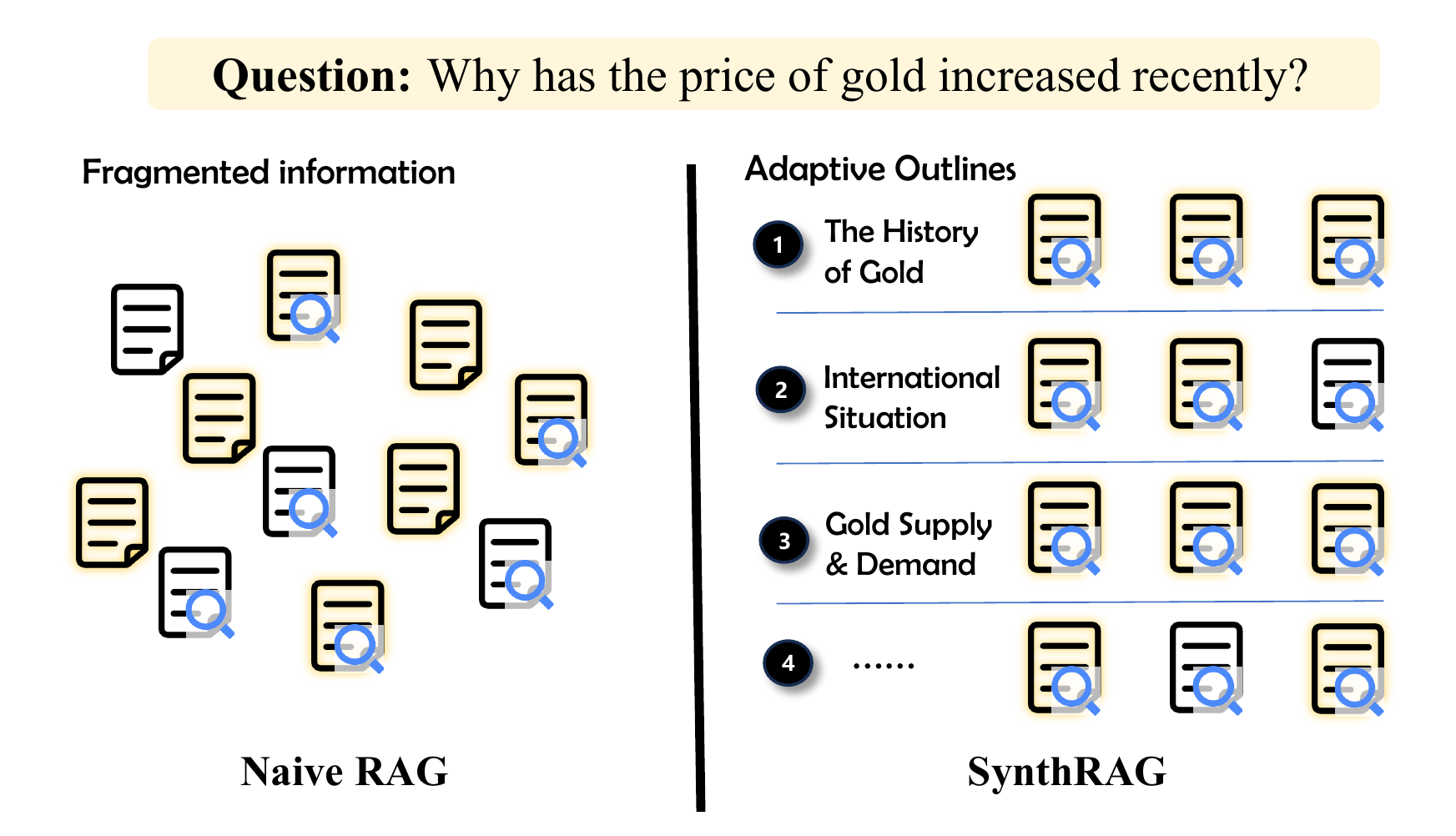}
  \caption{An illustration of naive RAG and SynthRAG. SynthRAG adeptly integrates and organizes vast, fragmented information, creating a comprehensive and systematic structure.}
  \label{fig:intro}
  \vspace{-10pt}
\end{figure}

% Existing 
Large Language Models (LLMs) excel at handling straightforward QA tasks where the ideal responses are factually concise. However, in cases that require synthesizing supporting references from diverse domains and producing logically coherent content, they often fall short~\cite{DBLP:journals/corr/abs-2311-09149}.
The reason might lie on the extra capabilities in exploration and integration, instead of solely focusing on generating fluent contexts.

As a response, Retrieval-Augmented Generation (RAG) models have become an effective approach for generating long-form answers by retrieving relevant information~\cite{gao2023retrieval}. 
The initial RAG method consists of simple indexing, retrieval, and generation processes; however, it fails to consider retrieval precision and generation quality~\cite{DBLP:journals/corr/abs-2312-10997}. 
Following studies improve these processes by employing pre-retrieval optimization, post-retrieval optimization, and iterative retrieval techniques~\cite{DBLP:journals/corr/abs-2002-08909,DBLP:conf/eacl/IzacardG21,DBLP:conf/icml/BorgeaudMHCRM0L22,DBLP:conf/emnlp/ShaoGSHDC23}. thereby increasing the accuracy and contextual relevance of generated answers. 
Although current RAGs can supplement LLMs with up-to-date information, they still lack the ability to identify the intricate features of such complex QA situations. 
In comparison to general answers, the generation is supposed to be both comprehensive and in-depth, ensuring coverage of potential domains and details.
Therefore, we first coin the task as\textit{ explanatory answer generation} and outline its notorious challenges as follows.

\textit{Challenges.~} 
First, the comprising components of the explanatory answer should be complete.
In order to answer the question in Fig.~\ref{fig:intro}, it is necessary to simultaneously consider topics such as ``The History of Gold'', ``International Situation'' and ``Gold Supply \& Demand'' to construct a thorough result.
The use of keywords and question embeddings as key clues in a naive RAG method tends to easily overlook relevant information that might not be explicitly mentioned in the question, resulting in insufficient information coverage and potential biased.
Second, it is significant to logically organize the content for the explanatory answer. 
Naive RAG methods usually retrieve fragmented information, leading to answers that merely aggregate isolated pieces of information. As shown in the illustration, while the information is relevant, it lacks structural organization.
Consequently, these responses frequently lack logical coherence and depth, as they fail to integrate the retrieved information into a coherent whole. 

There is a pressing need for approaches that can comprehensively retrieve information and employ systematic thinking to enhance the completeness and logic of the generated answers.
Gestalt psychology~\cite{kohler1967gestalt} highlights that human cognition is based on the integration of information into a unified whole, rather than simply accumulating fragmented data. 
This holistic approach is essential for problem-solving and knowledge construction. In the context of complex question answering, scattered information alone is insufficient to meet the user's needs for comprehensive and systematic responses.
Inspired by the theory, we focus on enhancing the capabilities of LLMs in effectively integrating and systematizing information. In line with this goal, we propose the SynthRAG framework, designed to enable LLM in comprehending and answering questions from a systematic perspective. 

Specifically, SynthRAG can incorporate existing RAG methods, making it a flexible and comprehensive solution. It consists of three main steps: adaptive outline generation, systematic information generation, and customized answer generation. In adaptive outline generation, SynthRAG learns outline instructions for different type of questions by analyzing high- and low-quality samples, then selects and adapts the appropriate outline based on the specific question to ensure comprehensive coverage and logical structure. During systematic information generation, SynthRAG retrieves and generates coherent detailed paragraphs for each subsection, resulting in comprehensive and logically structured responses. Finally, in customized answer generation, SynthRAG uses representative examples to help the model understand the problem globally, focusing on essential information to produce accurate and insightful answers. This method integrates diverse information while maintaining coherence and integrity, providing high-quality, in-depth responses that meet the user's requirements.
In summary, our contribution is three-fold:
% \vspace{-10pt}
\begin{itemize}
\item We systematically analyze the limitations of LLMs in handling questions that need comprehensive and in-depth responses, identifying deficiencies in information integration and answer depth.

\item We propose SynthRAG, a novel Retrieval-Augmented Generation framework that effectively integrates information fragments to provide a comprehensive and robust knowledge foundation for LLMs.

\item SynthRAG incorporates a holistic perspective that fosters high-level associative thinking and reasoning, addressing the critical issue of fragmented information linkage in traditional RAG models.

\item Through a series of human and LLM-based evaluations, our approach demonstrates significant advantages over existing methods. Results from the online deployment showed that the SynthRAG responses are well received. 

\end{itemize}

\section{Related Works}

In this paper, we explore the problem of knowledge integration for explanatory answer generation. Thus, we review prior work
% in three related fields,  i.e., retrieval augmented generation, explanatory answer generation, and generated response evaluation.
in retrieval augmented generation and explanatory answer generation.

\subsection{Retrieval-Augmented Generation}
Integration with external knowledge has become a prevalent strategy in QA tasks~\cite{DBLP:journals/corr/abs-2302-12813}. 
Models like REALM~\cite{DBLP:journals/corr/abs-2002-08909} employ dual-encoder and retrieval mechanisms to dynamically retrieve documents during answer generation. Advances such as the Fusion-in-Decoder~\cite{DBLP:conf/eacl/IzacardG21} further refine this approach by merging retrieved documents directly into the input layer, delivering more precise answers through an integrated retrieval-and-generation process. Meanwhile, Self-RAG~\cite{DBLP:conf/iclr/AsaiWWSH24} employs a self-reflection mechanism to optimize both retrieval and generation, showing notable efficacy in handling long text and open-domain QA tasks. Self-Reasoning~\cite{DBLP:journals/corr/abs-2407-19813} improves the generation quality and traceability~\cite{DBLP:journals/corr/abs-2311-03731} of retrieval-augmented language models by generating reasoning trajectories.
The Step-Back Prompting~\cite{DBLP:conf/iclr/ZhengMCCCLZ24} method improves the model's performance in complex reasoning tasks by guiding it to perform high-level abstraction and reasoning.
However, external knowledge is often fragmented, and these methods fail to systematize it. A holistic perspective is essential for problem-solving and knowledge construction. 
GraphRAG~\cite{DBLP:journals/corr/abs-2404-16130} provides a global perspective for LLMs by constructing a knowledge graph, enhancing the quality of responses effectively. 
Different from these work, we treat the retrieval step as a black box and focus on improving the generation quality given the query and retrieved passages. 
%Scattered information fragments cannot provide comprehensive answers. To improve LLMs in these scenarios, we need a method that effectively integrates and systematizes information for thorough, systematic, and in-depth responses.

% \subsection{Question Answering Systems}

\subsection{Explanatory Answer Generation}
Explanatory Answer Generation (EAG) stands apart from traditional QA system and article generation. 
Traditional QA systems~\cite{DBLP:conf/acl/SaxenaKG22,DBLP:journals/corr/abs-2311-09149,DBLP:journals/kbs/ChenZLLK22,DBLP:journals/apin/EtezadiS23,li2023lmeye,li2023multi} typically generate answers that consist of single entities, short phrases, or concise sentences. These responses are suitable for straightforward knowledge-based questions, where users seek concise information. However, such responses often fall short when dealing with more intricate inquiries that require a detailed exploration of the subject matter.
Long-form QA systems~\cite{DBLP:conf/acl/QinCJYLZLHDWXQL23,DBLP:conf/kdd/LiuLYXZDZDT23} have been developed to bridge this gap, but they often lack the depth and structured coherence needed for comprehensive explanations. 
FoRAG~\cite{DBLP:journals/corr/abs-2406-13779} refines long-form question answering through integrating an outline-enhanced generator and a fine-grained reinforcement learning framework.  Recently, some article generation work~\cite{DBLP:journals/corr/abs-2402-14207,DBLP:conf/emnlp/YangTPK22} has also begun to attempt generating logical text by first creating an outline and then writing the full text. However, these methods are often limited to specific professional fields and lack general applicability~\cite{DBLP:journals/corr/abs-2402-14207}. 

% Different from these work, EAG seeks to address this deficiency by not only extending the length of answers but also enhancing their quality and depth.

Unlike these studies, this research focuses on building a general QA system capable of providing comprehensive, systematic, and in-depth answers to user questions, effectively meeting user needs for thorough understanding.

% \subsection{Generated Response Evaluation}
% Supervised metrics like ROUGE and BLEU are widely recognized as inadequate for evaluating natural language generation outputs, particularly for open-ended writing tasks. Traditionally, evaluation has relied on rubric-based human assessment, which is costly and time consuming~\cite{fu2023gptscore}. Recent advancements in large language models (LLMs) have spurred interest in new evaluation paradigms that leverage these models for assessing text generated by LLMs~\cite{bai2024benchmarking,chang2024survey}. To effectively evaluate free-form text, a more customized and interactive framework is required.
% We utilize a multi-faceted evaluation approach to comprehensively assess the effectiveness of our model. This includes using GPT-4 for scoring based on specific criteria, reward model preference evaluation, assessment of the information content, and real-time feedback from online users. This holistic evaluation ensures a thorough understanding of the model's performance.

\section{METHODOLOGY}

In this study, we propose the SynthRAG framework, a novel systematic retrieval-augmented generation approach aimed at enhancing the performance of LLMs in explanatory answer generation tasks. 

\begin{figure*}[ht]
  \centering
  \includegraphics[width=\linewidth]{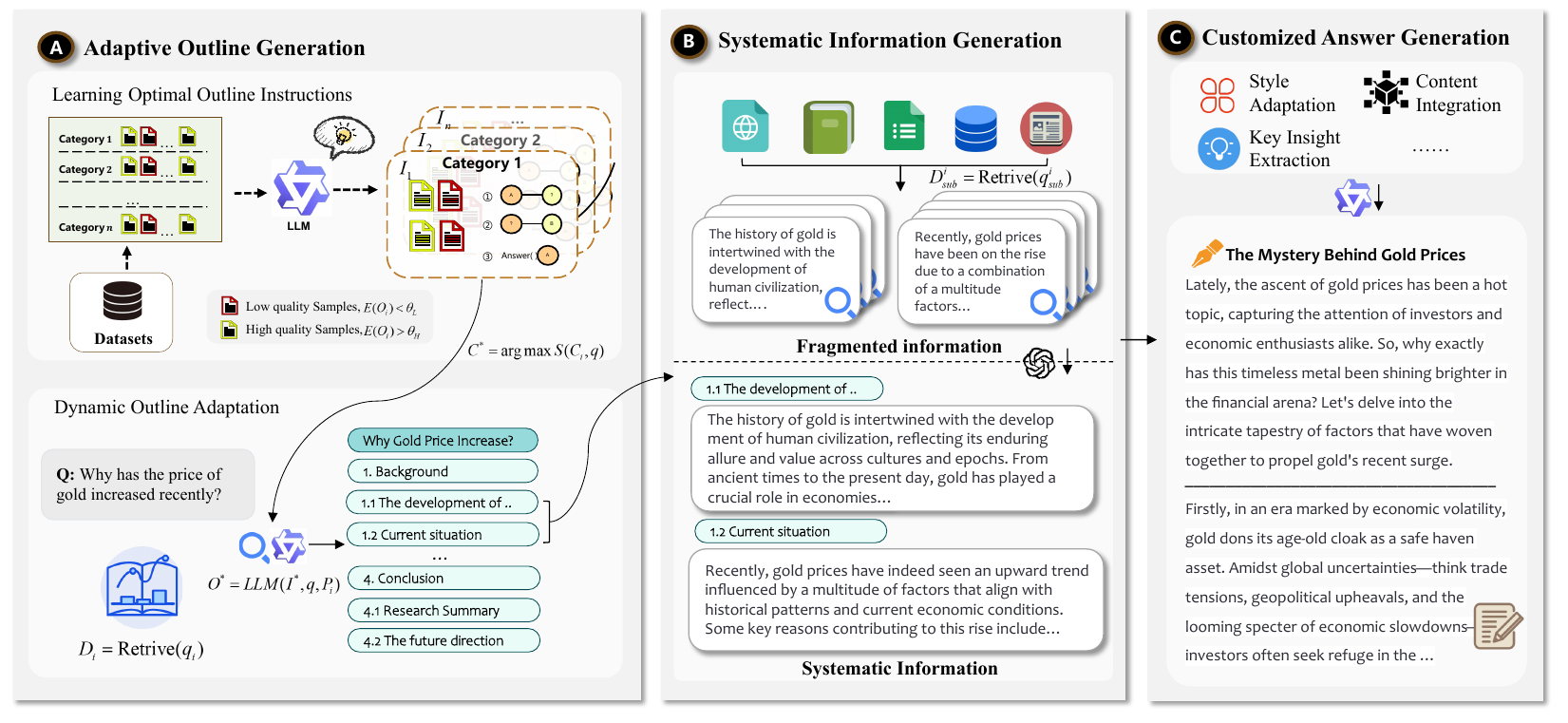}
  \caption{The SynthRAG framework consists of three primary components: adaptive outline generation, systematic chapter-level information generation, and customized answer generation. The integration of these components within the SynthRAG framework addresses the fragmentation and incompleteness issues of traditional RAG methods, providing a robust solution for explanatory answer generation tasks.
  }
  \label{fig:model}
  \vspace{-10pt}
\end{figure*}

\subsection{Problem Formulation}

The primary task of this research is to improve the performance of LLMs in explanatory answer generation scenarios. The main objective is to enable LLMs to generate coherent, comprehensive, and insightful answers by integrating information from diverse sources and presenting it in a structured manner.

Specifically, the task involves creating a system that, given a query \( q \), retrieves relevant external information \( D \) and generates an answer \( a \). The answer \( a \) should comprehensively and thoroughly address the query \( q \). Unlike traditional QA tasks that focus on knowledge-based or simple questions, this task emphasizes real-world user queries that are diverse and complex. These queries encompass various types, including knowledge-based, opinion-based, evaluative, and experiential questions, each demanding a high level of informational richness and answer quality.
This task requires addressing the limitations of naive RAG methods, which often struggle with fragmented and incomplete information synthesis. Our approach aims to provide a robust framework that improves the overall quality and depth of the answers generated by LLM.

\subsection{SynthRAG Framework}
\label{sec:overall}

%The SynthRAG framework is designed to incorporate systematic retrieval and generation techniques to enhance the performance of LLMs in explanatory answer generation tasks. 
As shown in Fig.~\ref{fig:model}, the SynthRAG framework consists of three primary components: adaptive outline generation, systematic information generation, and customized answer generation.
Firstly, the adaptive outline generation component leverages historical data to learn optimal outline structures for different question types. By analyzing both high- and low-quality examples, the system identifies the most effective outlines and dynamically adjusts them to ensure comprehensive coverage of key information and a coherent logical flow.
Secondly, the systematic information generation component deepens the retrieval and generation process, creating detailed and coherent information for each outline subsection. This approach ensures smooth information transitions and logical consistency, building a holistic view from macro to micro levels.
Finally, the customized answer generation component employs representative examples to guide the LLM in understanding the question globally. This phase refines the generated information, ensuring that the final answer is insightful.

\subsubsection{Adaptive Outline Generation}
Different types of questions require varied response strategies. The adaptive outline generation component ensures comprehensive coverage and a logical structure in the generated content through two key stages: learning optimal outline structures from historical data, and dynamically adapting these structures to suit specific questions.

% The adaptive outline generation component ensures comprehensive coverage and logical structure of the generated content. This involves two key stages: learning optimal outline structures from historical data and dynamically adapting these structures to fit specific question.

\paragraph{Learning Optimal Outline Structures}

The key step in adaptive outline generation is to learn the optimal outline structures from historical data for different question types. This involves analyzing high-quality and low-quality samples to identify patterns and structures that lead to effective and comprehensive answers.

Let \( Q = \{q_1,q_2,...,q_n\} \) denote the set of queries and \( A = \{a_1,a_2,...,a_n\} \) denote the set of historical answers. In this work, we use Zhihu-KOL~\footnote{\url{https://huggingface.co/datasets/wangrui6/Zhihu-KOL}} as historical data, which contains 1 million real QA pairs. To learn the optimal structure of the responses, we first calculate the effectiveness score \( E(a_i) \) for each answer based on predefined criteria such as coherence, completeness and user satisfaction. The effectiveness score is given by:
\begin{equation}
E(a_i) = \alpha \cdot H(q_i, a_i) + \beta \cdot P(a_i) + \gamma \cdot U(a_i),
\end{equation}

where the coherence score \( H(a_i) \) measures the semantic similarity between the query and the answer by calculating the cosine similarity between their embeddings. %This process ensures that the content is not only relevant but also logically aligned with the query. 
The completeness score \( P(a_i) \) is based on the length of the text, as overly short responses make it difficult to learn a consistent and repeatable answer structure. The score is then normalized to indicate how thoroughly the response addresses the required information.
The user satisfaction score \( U(a_i) \) is a crucial metric derived from the number of likes or positive feedback on online platforms, reflecting the practical utility and acceptance of the answer among users.
All these scores are normalized to ensure consistency and comparability. \( \alpha \), \( \beta \), and \( \gamma \) are weighting factors determined through empirical analysis~\footnote{In this work, we use the weights $\alpha = 0.3$, $\beta = 0.1$, $\gamma = 0.6$. We prioritize user preferences and aim to eliminate overly brief answers, as these may be opportunistic and lack reproducibility}.

Consequently, we utilize GTE~\cite{DBLP:journals/corr/abs-2308-03281} to encode the text and employ the unsupervised clustering method K-means~\cite{MacQueen1967SomeMF} to categorize these historical samples into $k$ distinct clusters.
Let \( \{C_1, C_2, \ldots, C_k\} \) be the set of question type clusters obtained through K-means clustering. For each cluster \( C_j \in \{C_1, C_2,...,C_k \}\), we define the set of high-quality samples \( H_j \) and low-quality samples \( L_j \) as follows:
\begin{equation}
H_j = \{ a_i \mid E(a_i) \geq \theta_H \},
\end{equation}
\begin{equation}
L_j = \{ a_i \mid E(a_i) < \theta_L \},
\end{equation}
where \( \theta_H \) and \( \theta_L \) are are dynamically selected for each question category. These thresholds are set to identify the top 5\% and bottom 5\% of the answer samples as positive and negative samples, respectively. 

Identifying \( H_j \) and \( L_j \) within each cluster, we enable the LLM to actively learn from these specific historical instances. The LLM analyzes the characteristics of high-quality and low-quality samples, summarizing them to distill detailed outline instructions \( I_j \) for each query type:
\begin{equation}
I_j = \text{LLM}(H_j, L_j).
\end{equation}

%Fig~\ref{fig:instruct_generate_prompt} and \ref{fig:instruct_sample} presents 
The detailed prompt and samples of the generated $I_j$ can be found on our code repository.
These comprehensive outline instructions \( I_j \) guide the adaptive outline generation process, ensuring that the model leverages the insights from historical data to construct effective and coherent outlines for different question types. 

\paragraph{Dynamic Outline Adaptation}

Once the optimal outline instructions for different question types are learned, the next step is to dynamically adapt these outline instructions to fit specific user queries. 
Specifically, for a given question \( q_i \), we calculate the cosine similarity score \( S(C_j, q_i) \) for each cluster \( C_j \), and select the cluster \( C^*_i \) with the highest score to the question \( q_i \):
\begin{equation}
C^*_i = \arg \max_{C_j} S(C_j, q_i),
\end{equation}
where similarity score \( S(C_j, q_i) \) measures how closely the characteristics of the cluster \( C_j \) match the query \( q_i \), typically calculated using cosine similarity.

After selecting the optimal outline instructions \( I^*_i \) , the system dynamically adapts it to the specific outlines. This involves adjusting the depth and breadth of each section based on the query's complexity and the retrieved information.
\begin{equation}
    O^*_i  = \text{LLM}(I^*_i,q_i,D_i),
\end{equation}
where  \( I^*_i \) is the outline instruction of \( C^*_i \) 
%(Fig.~\ref{fig:instruct_sample} shows an example of outline instruction).
$D_i$ is retrieved documents and $O^*_i$ is the specific outline output. \( O^*_i \) is a sequence of sections and subsections,\( O^*_i = \{ S_{i1}, S_{i2}, \ldots, S_{in} \}, \) where \( S_{ij} \) represents the \( j \)-th section of the outline for query \( q_i \).

This approach enables the system to dynamically adjust outline instructions to suit the specific characteristics of each query, using insights from historical data for structured and comprehensive responses.

% This approach ensures that the system dynamically adapts the outline instructions to fit the specific characteristics of the query, leveraging the insights from historical data to provide a structured and comprehensive response.
% The adaptive outline generation process ensures that the generated content is well-structured, logically coherent, and comprehensive, addressing the specific needs of each query while maintaining high standards of quality and relevance.

\subsubsection{Systematic Information Generation}
\label{sec:Chapter-Level Information Generation}

The systematic information generation component of the SynthRAG framework is designed to ensure that each section of the adaptive outline is populated with detailed, coherent, and contextually relevant information. This process involves the following key steps: hierarchical information retrieval, parallel content generation, and logical consistency enforcement.

\paragraph{Hierarchical Information Retrieval}
% The first step is to retrieve the relevant information hierarchically based on the adaptive outline.
Given an outline \( O^*_i \) with sections \( \{ S_{i1}, S_{i2}, \ldots, S_{in} \} \), we perform targeted retrieval for each section \( S_{ij} \).
Hence, \( S_{ij} \) can be regarded as the sub-query derived from the main query \( q_i \) that focuses on the specific topic of \( S_{ij} \). The information retrieval process involves fetching relevant documents \( D_{S_{ij}} \) from a knowledge base or external sources. %:
% \begin{equation}
% D_{S_{ij}} = \text{Retrieve}(S_{ij}),
% \end{equation}
% where \( \text{Retrieve} \) is the retrieval function that returns a set of documents relevant to the sub-query. 
Note that the retrieval method here can be replaced by any mainstream RAG method. In this work, a basic search engine retrieval method is employed.

\paragraph{Parallel Content Generation}
\label{sec:Parallel}
Upon retrieving the relevant documents for each section \( S_{ij} \), the next phase is to generate detailed content for each section in parallel. Our objective is to ensure comprehensiveness and coherence while optimizing for cost-effectiveness and time efficiency. 

Let \( G_{ij} \) represent the generated content for section \( S_{ij} \). LLM takes the outline \( O^*_i \), the specific sub-query \( S_{ij} \), and the retrieved documents \( D_{S_{ij}} \) as inputs to produce the section content:
\begin{equation}
G_{ij} = \text{LLM}(O^*_i, q_i, S_{ij}, D_{S_{ij}}).
\end{equation}

This method leverages the structured outline to maintain the overall coherence and logical flow of the final response. Under the guidance of a unified outline, we generate the content for each section in parallel. It ensures that each section is coherent and logically consistent, despite being generated simultaneously. The generated sections are then merged to form a complete, comprehensive answer.

\subsubsection{Customized Answer Generation}
SynthRAG enhances systematically generated outputs by utilizing a set of complete, high-quality historical answers of the same type. These exemplary responses are provided to the LLM as references, allowing it to gain a more holistic understanding of the query. During the refinement process, the LLM uses these high-quality answers to guide content condensation, focusing on key insights while eliminating redundancy. In addition, it adapts the language, tone and structure to align with the style and expectations set by these examples, ensuring that the final output is informative and adheres to the required formatting and readability standards. By drawing on these references, the LLM significantly improves the systematically generated content, delivering a customized answer that is well-structured, relevant, and coherent, and effectively addresses the specific needs of the query.

\begin{table*}[]
\resizebox{0.95\textwidth}{!}{%
\begin{tabular}{l|cccccc|c}
\hline
\multirow{2}{*}{\textbf{Model}} & \multicolumn{6}{c|}{\textbf{LLM-base Score}}                                                                    & \textbf{Information Score} \\ \cline{2-8} 
                                & \textbf{Overall} & \textbf{Fluency} & \textbf{Relevance} & \textbf{Logic} & \textbf{Reference} & \textbf{Depth} & \textbf{Overall (\%)}                   \\ \hline
ChatGPT                         & 3.76             & 3.92             & 4.34               & 4.00            & 3.50                & 3.24           & \textbackslash{}                   \\
GPT-4                           & 3.82             & 3.90              & 4.40                & 3.90            & 3.64               & 3.28           & \textbackslash{}                   \\
Qwen-Max                        & 4.06             & \textbf{4.00}              & 4.58               & 4.28           & 3.96               & 3.88           & 63.91                              \\
RAG                             & 4.12             & 3.94              & \underline{4.70}                & \underline{4.22}           & 4.04               & 3.86           & 66.43                              \\ 
OutlineRAG        & \underline{4.34}             & \textbf{4.00}              & 4.60                & \underline{4.22}           & \underline{4.16}               & \underline{4.32}           & \underline{83.01}                              \\
\hline
\textbf{SynthRAG}                        & \textbf{4.52}             & \underline{3.96}             & \textbf{4.74}               & \textbf{4.30 }           & \textbf{4.40}                & \textbf{4.54}           & \textbf{85.76}                              \\
\quad w/o generation         & 3.86             & \textbf{4.00}              & 4.18               & 3.76           & 3.78               & 3.58           & 74.44       
\\
\quad w/o customized          & 4.42            & \textbf{3.96}              & 4.66               & 4.18           & 4.14               & 4.28           & 85.64     
\\ \hline
\end{tabular}%
}
\caption{Experimental results based on model evaluation and information content assessment. The LLM-base score ranges from 0 to 5, with higher scores indicating better performance. Detailed scoring criteria are provided in Table~\ref{table:Evaluation_Metrics}. The Information Score ranges from 0 to 100, with higher scores indicating more comprehensive and richer information content in the answers.}
\label{tab:my-table}
\end{table*}

\section{Experiment}
In this section, we conduct extensive experiments to validate the effectiveness of our framework.
 
% \subsection{Question Selection}
\subsection{Experimental Setup}
\subsubsection{Question Selection}
SynthRAG is designed to provide comprehensive answers in explanatory answer generation scenarios. To avoid issues such as data leakage, we randomly selected questions from the daily hot list on Zhihu\footnote{\url{https://www.zhihu.com}, one of the largest Chinese Q\&A communities.} from April 2024 to May 2024. These questions cover the latest social dynamics, news events, and public opinion information, making the categories complex and diverse. Most of these questions require a good reserve of relevant knowledge and experience to answer effectively. A total of 50 questions were selected for subsequent experimental evaluation.

\subsubsection{Implementation Details}

We use \texttt{Qwen-Max}~\footnote{https://qwenlm.github.io/blog/qwen-max-0428/} as our base LLM.  We set the temperature at 1.0 and the top\_p at 0.9 for all experiments. Text is encoded using GTE~\cite{DBLP:journals/corr/abs-2308-03281}, and setting the number of clusters $k$ to 100. 
%The original history QA data are from https://huggingface.co/datasets/wangrui6/Zhihu-KOL, which contains 1M QA pair.
For reported results, the \ourmodel is grounded on the Quark search API with top 5 chunks, although the proposed methods is compatible with all other RAG methods. 

\subsubsection{Compared Methods}
As prior works focus on shorter answers or use different setups, they are difficult to compare directly. 
Instead, we use the following baselines in our analysis: Direct Generation, where the LLM is prompted to directly produce answers; RAG, which searches by topic and integrates the retrieved results with the topic to generate responses; and OutlineRAG, which initially creates an outline, then searches for additional information using section titles to formulate the answer.

\subsection{Evaluation Metrics}

To comprehensively evaluate the effectiveness of our answers, we designed four types of evaluation metrics:

\subsubsection{LLM-Based Evaluation} Following the recent trend of using LLMs to judge the quality of output text (especially in reference-free evaluation settings), we employed GPT-4 to assess the quality of generated answers. The evaluation is based on five different criteria. We then ask the evaluation model to rate each answer on a 5-point Likert scale or to make pairwise comparisons between the ideas generated by different models. Detailed human-crafted criteria used to guide the evaluation are provided in Table~\ref{table:Evaluation_Metrics}.

\subsubsection{Reward Model-Based Ordinal Evaluation} 
\label{sec:reward}
To effectively evaluate the relative merits of two answers, we categorized responses into high and low upvote pairs based on publicly available QA pairs from Zhihu~\footnote{\url{https://huggingface.co/datasets/wangrui6/Zhihu-KOL}}, and trained a reward model to discern human preferences between the two. This model ranks answers by predicting which one is more likely to be favored by humans. Specifically, we constructed a large set of answer preference samples \((Q, A_1, A_2)\), where \(Q\) represents a question, and \(A_1\) and \(A_2\) are two answers to that question, adhering to the following conditions: i) Both \(A_1\) and \(A_2\) are responses to the same question \(Q\). ii) The number of user upvotes for \(A_1\) exceeds that of \(A_2\). iii) The posting time of \(A_1\) is not earlier than that of \(A_2\).
% \begin{enumerate}
%     \item Both \(A_1\) and \(A_2\) are responses to the same question \(Q\).
%     \item The number of user upvotes for \(A_1\) exceeds that of \(A_2\).
%     \item The posting time of \(A_1\) is not earlier than that of \(A_2\).
% \end{enumerate}
By doing so, we mitigated biases arising from differences in questions and answer exposure times. We extracted a total of 100k answer preference pairs for training our reward model, with an additional 5k pairs reserved for evaluating its performance. We fine-tuned the Qwen1.5-1.8B~\cite{qwen1.5} model using Llama-Factory~\cite{zheng2024llamafactory} on an NVIDIA A100 80GB GPU. The ordinal accuracy of the reward model on the test set reached 84.3\%.
This model helps to rank the answers by predicting which one is more likely to be preferred by humans.

% \paragraph{Information Content Evaluation} To further assess the information content of the model outputs, we designed an evaluation method based on question answering. We concatenated answers generated by different models as the context and let the LLM generate specific questions based on this context. 
% More details can be found in Appendix~\ref{sec:info_eval}

\subsubsection{Information Content Evaluation}
\label{sec:info_eval}
To better assess the informational richness of model outputs, we propose an evaluation framework based on factual question answering. This approach aggregates answers from multiple models to create a comprehensive context, which an LLM then uses to generate specific questions for information assessment. In the evaluation phase, each individual answer is treated as a standalone context, prompting the LLM to respond to these pre-generated questions using only the given response. The precision of these answers acts as a proxy for the information density of each model's output.

\paragraph{Test QA Pair Generation.} 
For a given query $Q$, let $A_i$ represent the answer generated by model $i$, where $i \in [1, n]$ and $n$ is the total number of models. The aggregate context $\text{Context}_{\text{all}}$ is formed by concatenating all the answers $A_i$:
\begin{equation}
\text{Context}_{\text{all}} = \text{Concat}(A_1, A_2, \ldots, A_n),
\end{equation}
capturing the diverse perspectives of all models regarding the query.

Using $\text{Context}_{\text{all}}$, the LLM generates question-answer pairs $(q, a)$ designed to be answerable with the aggregated context:
\begin{equation}
[(q_1, a_1), \dots, (q_n, a_n)] = \text{LLM}(\text{Context}_{\text{all}}).
\end{equation}
These pairs ensure that the generated questions can be answered using the combined information from all models.

\paragraph{QA Evaluation.}
The generated question-answer pairs are then used to evaluate each model individually. For model \( i \), its specific context \( A_i \) is provided to the LLM to answer the previously generated questions. The precision of the LLM's answers, based on \( A_i \), reflects the information density of that model’s response. 

The Information Score, ranging from 0 to 100, is computed based on the accuracy of the LLM's answers, with higher scores indicating more informative and comprehensive content. This precision-based evaluation provides a quantifiable metric for assessing the information richness of each model, ensuring an objective and thorough evaluation.

% \subsubsection{Online Human Evaluation} We registered virtual Zhihu accounts and posted the generated answers on the Zhihu platform. We collected real user feedback data from the platform to evaluate the effectiveness of the answers. This includes metrics such as upvotes, comments, and shares to gauge the user engagement and satisfaction with the answers.

\subsection{Main Results}
\textbf{LLM-Based Evaluation.}
The LLM-based evaluation measures the performance of different models across several dimensions, including overall quality, fluency, relevance, logic, reference value, and depth. The scores for each dimension range from 0 to 5, with higher scores indicating better performance. 
%The specific rating criteria employ the Likert scale, which can be found in Table~\ref{table:Evaluation_Metrics}. 
The results for each model are summarized in Table~\ref{tab:my-table}.
SynthRAG scored the highest overall at 4.52, surpassing baseline models ChatGPT and GPT-4, which registered scores of 3.76 and 3.82, respectively. It also achieved comparable fluency scores, at 3.96, similar to Qwen-Max and RAG at 4.00 and 3.94. Moreover, SynthRAG's relevance score of 4.74 exceeded those of ChatGPT (4.34) and GPT-4 (4.40), indicating its superior alignment with query contexts. Its logic score of 4.30, alongside reference value and depth scores of 4.40 and 4.54, respectively, further demonstrate its ability to generate valuable and insightful content, outstripping other models.
Overall, the LLM-based evaluation demonstrates that SynthRAG significantly outperforms existing models in terms of overall quality, relevance, logic, reference value, and depth, while maintaining competitive fluency.

\begin{figure*}[ht]
  \centering
  \includegraphics[width=\linewidth]{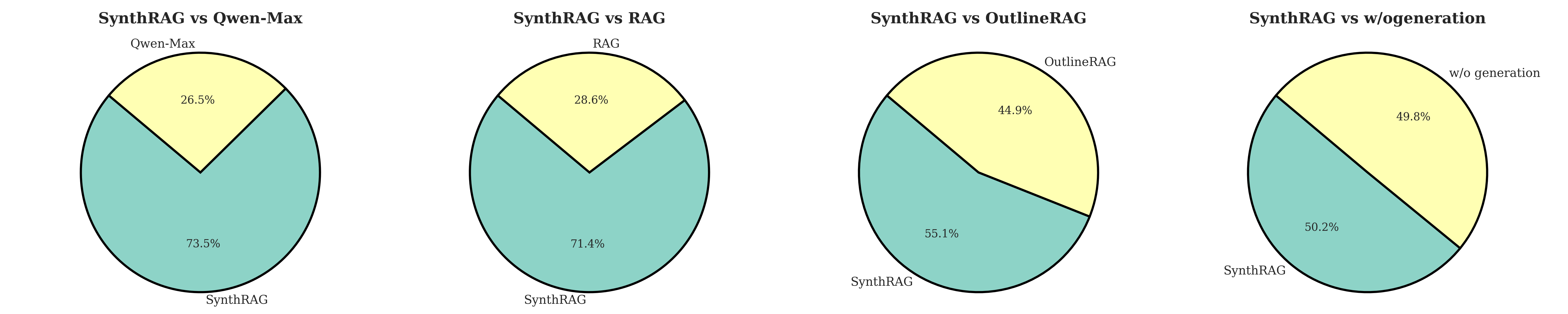}
  \caption{Results of reward model evaluation. The reward model-based ordinal evaluation demonstrates the superiority of SynthRAG over baseline models and ablation version. SynthRAG consistently achieves higher user preference scores, indicating its effectiveness in generating answers that align more closely with user tastes.}
  \label{fig:reward_model}
\end{figure*}

\textbf{Information Content Evaluation.}
We used generated QA pairs (see details in section~\ref{sec:info_eval}) to evaluate the information content of each model's answers. As presented in Table~\ref{tab:my-table},  
RAG scored 66.43, slightly outperforming Qwen-Max. This score reflects RAG's capacity to incorporate retrieved information into its answers. However, it remains significantly lower than OutlineRAG, indicating that a single retrieval and generation cycle may struggle to produce comprehensive responses.
SynthRAG achieved the highest information score of 85.76, markedly surpassing other models. Comparisons with OutlineRAG reveal that providing a specific outline instruction not only enhances the comprehensiveness but also the target-specificity of the generated content, aiding in the creation of more informative responses. 
% This result underscores SynthRAG's superior ability to generate answers that are both extensive and rich in content, establishing it as the foremost model in terms of information comprehensiveness.
This result highlights SynthRAG's effectiveness in generating answers with extensive and rich content, making it the leading model in terms of information comprehensiveness.

%Qwen-Max, with an Information Score of 63.91, demonstrated a good level of information richness but fell short compared to SynthRAG. This indicates that while Qwen-Max can generate informative answers, it lacks the depth and comprehensiveness provided by SynthRAG. 

\subsection{Ablation Studies}
To evaluate the efficacy of the individual components of the model, we conducted ablation studies.

Initially, we removed the instruction guidance component (OutineRAG), the LLM-base score and the information score were both negatively affected. The LLM-base score decreased to 4.34, and the information score dropped to 83.01. Although this version of SynthRAG still performed well, the reduction in scores underscores the importance of instruction guidance in enhancing the overall quality and richness of generated information.

Similarly, omitting the systematic information generation component (w/o generation) resulted in a significant decline in both evaluation metrics. In this ablation setup, the systematic information generation module was excluded, and the LLM generated answers directly under the guidance of an outline, relying solely on retrieval. The LLM-base score fell to 3.86, and the information score plummeted to 74.44. Furthermore, this outcome's LLM score was even lower than the original Qwen-Max performance. This could be attributed to the fact that, while an outline was provided, it was unable to generate sufficiently comprehensive responses in a single retrieval and generation cycle. As a result, the answers produced under the guidance of the outline were overly generalized and vague, leading to a significant decrease in quality. This sharp decline underscores the critical role of the systematic information generation component in ensuring detailed and comprehensive answers.

Finally, we removed the customized answer generation component, which refines the output based on high-quality historical answers. The LLM-base score showed a slight decline from the full SynthRAG model, dropping to 4.42, and the information score fell to 85.64. Although the scores remained relatively high, this decrease suggests that the customized answer generation plays a important role in improving the final answer quality by refining the structure, tone, and relevance of the output. 

% These findings highlight that the performance achieved by the full SynthRAG framework is due to its superior ability to integrate diverse pieces of information and present them coherently and in detail. This capability is essential for addressing complex questions that require extensive knowledge and detailed explanations.

\subsection{Reward Model-Based Ordinal Evaluation}
To corroborate the effectiveness of our proposed model, we utilized authentic Zhihu preference data to train a reward model. Details of the training procedure are delineated in section~\ref{sec:reward}. The trained reward model exhibited proficiency in distinguishing answers that were more appealing to real users.
We performed pairwise assessments contrasting responses from our model against those from baseline and ablation models for the same question. The outcomes are shown in Fig.~\ref{fig:reward_model}. Our findings indicated that SynthRAG consistently outperformed baseline and ablation models in pairwise comparisons, aligning more closely with user preferences. This advantage stems from SynthRAG's ability to leverage insights from high-quality responses, enabling it to generate answers that resonate with users. %Notably, SynthRAG's performance remained strong even without the generation component, highlighting the importance of learned experience from high-quality data in improving model output.

% Our findings indicated that SynthRAG surpassed baseline and ablation models in pairwise comparisons, consistently ranking higher for the majority of questions. This suggests that the answers generated by SynthRAG are more in tune with actual user preferences. This superiority is largely due to SynthRAG's capability to harness experience gleaned from a massive collection of high-quality responses, facilitating the production of answers that are more attuned to user tastes.
% Comparing SynthRAG with SynthRAG w/o generation, it was observed that when guided by learned experiences, both models produced preference outcomes that were remarkably close. This further validates the significance of experience gained from analyzing large volumes of high-quality responses in enhancing model performance.

\subsection{Analysis of Computational Cost}
While SynthRAG introduces additional components, such as adaptive outline generation and systematic information retrieval, the overall computational cost remains manageable due to its parallel content generation mechanism.

We evaluated SynthRAG's computational cost compared to baseline models and found that, while it involves extra steps, the system is optimized to mitigate overhead. Specifically, SynthRAG requires an average of 15.6 retrievals and 16.6 LLM api calls for generation (most of which occur in parallel) . On average, it takes about 2 to 3 minutes to generate an explanatory answer.
Although this generation time makes SynthRAG less suitable for real-time, on-demand question answering, it is acceptable for more complex, explainability-focused scenarios. In many cases, answers can also be generated asynchronously, making the framework practical for high-depth QA tasks.

\section{Online application and evaluation}

\begin{figure}[]
  \centering
  \includegraphics[width=\linewidth]{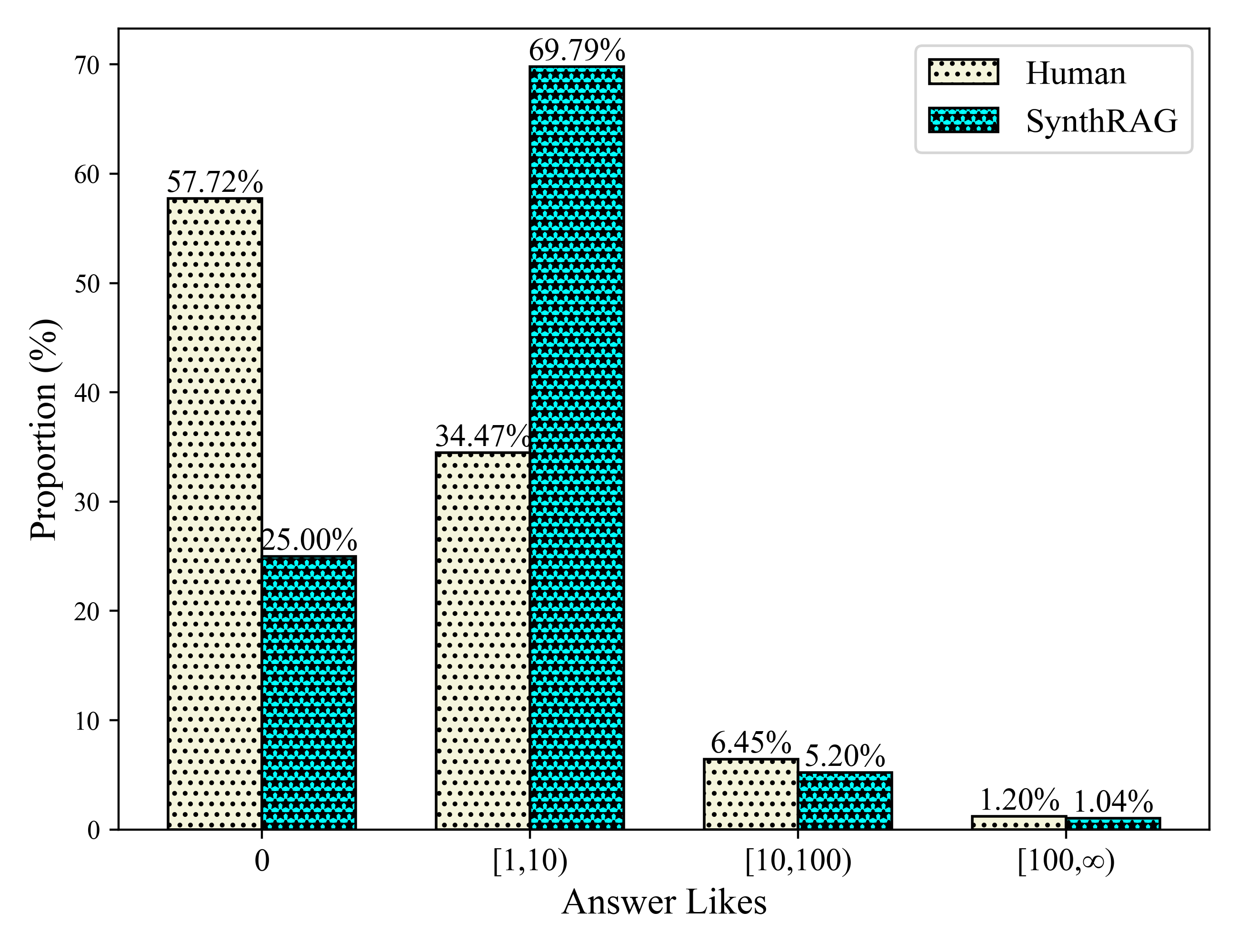}
  \caption{Comparison of Like Distributions for Human Answers and SynthRAG Generated Answers on Zhihu.}
  \label{fig:likes}
  \vspace{-2pt}
\end{figure}

To validate user preferences regarding our model's performance under real-world applications, we established a new account on the Zhihu platform. This account was dedicated to posting responses generated by our model. An automated system was implemented to select trending questions on Zhihu daily, generate content autonomously, and post these responses without human intervention. We periodically collected upvote metrics from real users to assess engagement.

During the two-month period from April to May 2024, we published 96 responses. These garnered an average of 5.73 upvotes per answer, with one response accumulating over 250 upvotes. The average upvote count exceeded 79.8\% of human Zhihu content creators~\footnote{Since many users have never answered a question, we ensured fairness in our comparison by focusing solely on Zhihu content creators, defined as users who have actively responded to questions. Our comparative data is based on the statistical analysis of 4.8 million Zhihu users.}, and our highest-rated answer ranked in the top 2\% of all upvoted responses on Zhihu.
Fig.~\ref{fig:likes} shows like distributions for human answers and SynthRAG generated answers on Zhihu. Further analysis revealed that only 25\% of SynthRAG's answers received no upvotes, compared to 57.74\% for human users. Moreover, 69.79\% of SynthRAG's answers received 1-10 upvotes, significantly higher than the 34.47\% for human answers. These statistics indicate that SynthRAG's responses generally met user needs and gained recognition. However, we observed that human users had a slightly higher proportion of highly upvoted answers, with 1.25\% higher in the 10-100 upvote range and 0.16\% higher in the 100+ upvote range. This suggests that human expertise in specific domains, often based on unique experiences or insider knowledge, remains valuable for generating top-tier content.
% High-upvote human responses often displayed in-depth domain experience and unique insights, revealing information that is difficult to find online, highlighting a distinct advantage of human-generated content.

\begin{figure}[]
  \centering
  \includegraphics[width=\linewidth]{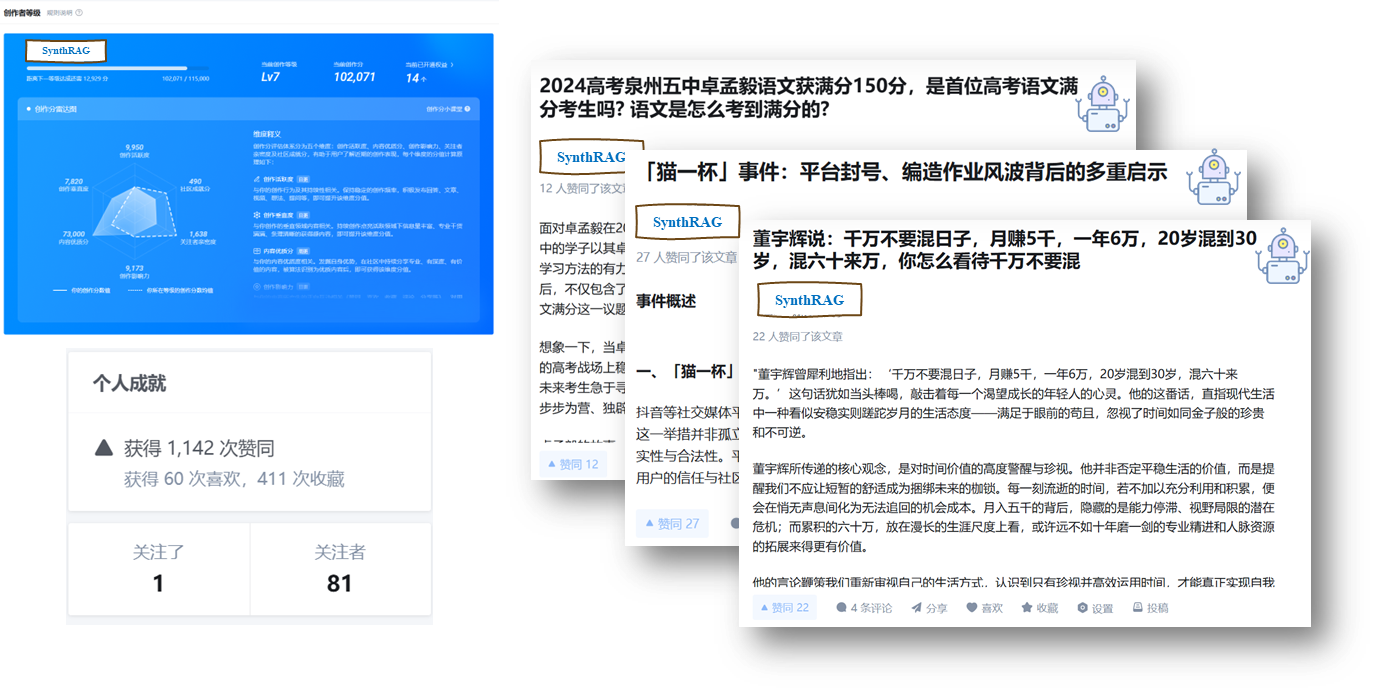}
  \caption{Various metrics such as upvotes, engagement, and follower growth of our zhihu bot account.}
  \label{fig:zhihu}
\end{figure}

As shown in Fig.~\ref{fig:zhihu}, the account gained 81 followers, placing it in the top 3.5\% of Zhihu users by follower count. This metric also reflects the user preference for our generated content. 
Notably, all above result was achieved without any initial follower base or external promotion, demonstrating the effectiveness and practicality of SynthRAG.
Examples of answers generated by SynthRAG can be found in Appendix~\ref{sec:examples}. These results demonstrate the practical efficacy of our model and its potential for real-world application and user engagement.

\section{DEPLOYMENT-CONTINUOUS HUMAN FEEDBACK}
\begin{figure}[]
    \centering
    \includegraphics[width=0.9\linewidth]{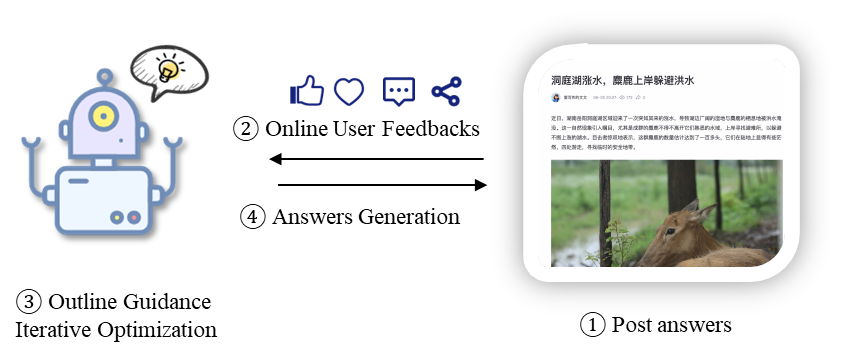}
    \caption{Illustration of the Iterative Feedback Process. User interaction data is then utilized to iteratively refine and enhance SynthRAG's responses, ensuring they meet users' preferences over time.}
    \label{fig:enter-label}
\end{figure}

Over time, the distribution of question types preferred by online users tends to diverge from offline data. Consequently, continual learning from user feedback is crucial for maintaining the advanced status of QA models.
SynthRAG demonstrates strong continuous optimization capabilities through the integration of online user feedback. The framework’s adaptive learning mechanism allows it to dynamically incorporate new information and feedback. 
This capability is particularly important in handling evolving queries, where the context and specifics can change over time. Fig.~\ref{fig:enter-label} illustrates of the iterative feedback process. By collecting both positive and negative real-world interaction samples, SynthRAG iteratively refines and expands its guidance library, thereby improving the quality and relevance of its responses. 

Table~\ref{tab:ongoing} displays the performance comparison between the initial model, which was learned from offline data, and the model after one iteration of online feedback. The original model achieved an LLM score of 4.52 and a reward model win rate of 41.6\%. In contrast, the SynthRAG-refinement model demonstrates enhanced performance, achieving an LLM score of 4.60 and a significantly increased RM win rate of 58.40\%.  
Continuous human feedback is vital to this iterative improvement process. User interactions, evidenced by metrics such as upvotes, comments, and shares, provide insights into the effectiveness of SynthRAG’s responses. SynthRAG utilizes this data to regularly update its guidelines, ensuring that its responses align with evolving user expectations and preferences.
This continuous refinement process not only improves response quality but also fosters greater user trust and engagement, establishing SynthRAG as an effective and user-centric AI-driven QA system.

\begin{table}[h]
% \resizebox{0.7\textwidth}{!}{%
\small
\begin{tabular}{l|c|c}
\hline
\textbf{Model} & \textbf{LLM Score} & \textbf{RM Win Rate (\%)} \\ \hline
SynthRAG-orginal        & 4.52             & 41.60                    \\
SynthRAG-refinement & \textbf{4.60} & \textbf{58.40}                      \\ \hline
\end{tabular}%
% }

% SynthRAG-Original represents the model based on initial offline Q\&A pairs, while SynthRAG-Refinement, optimized based on online feedback, demonstrates enhanced performance.
\caption{Performance metrics for two versions of SynthRAG.}
\label{tab:ongoing}
 \vspace{-10pt}
\end{table}

\section{Conclusion and Future Work}

The novel SynthRAG framework significantly enhances the performance of LLMs in explanatory answer generation tasks by integrating diverse information sources coherently and comprehensively. By leveraging adaptive outline generation, systematic information generation, and customized answer generation, SynthRAG improves the quality, depth, and coherence of responses. Extensive evaluations, including LLM-based, reward model-based, and online human assessments, demonstrate its efficacy and practical applicability, as evidenced by positive user feedback and high engagement on platforms like Zhihu. SynthRAG marks a significant step towards more insightful and user-centric AI-driven QA systems. In the future, we will focus on enhancing the model to user-specific adaptability. This involves tailoring responses to individual users, thereby achieving a higher degree of personalization answers.

\bibliographystyle{ACM-Reference-Format}
\bibliography{sample-base}

\appendix

% \section{Examples of SynthRAG Generated Answers}
% \label{sec:examples}

% %To illustrate the effectiveness of the SynthRAG framework, we present examples of generated answers and compare them with those from baseline and ablation models.
% Table~\ref{table:example} presents examples from the SynthRAG model, illustrating its adeptness at synthesizing diverse sources to produce structured, relevant answers. This makes SynthRAG particularly effective in scenarios requiring detailed and contextually nuanced responses.

% Figure~\ref{table:example} showcases specific instances of answers generated by SynthRAG. 
% These examples highlight SynthRAG's enhanced ability to address complex questions by integrating diverse information sources and presenting them in a structured manner. By doing so, SynthRAG ensures that the generated answers are not only comprehensive but also tailored to the specific context of the questions.

% These examples underscore the practical benefits and superior performance of the SynthRAG framework in generating high-quality answers. The framework's ability to synthesize information from multiple sources and its systematic approach to answer generation set it apart from traditional models. This makes SynthRAG particularly effective in scenarios requiring detailed and contextually nuanced responses.

\begin{table*}[]
\centering \small
\begin{tabular}{@{}p{3cm}p{0.5cm}p{10cm}@{}}
\toprule
\textbf{Metric} & \textbf{Score} & \textbf{Description} \\ 
\midrule

\multirow{5}{*}{Fluency} & 1 & Content is disorganized, expressions are awkward, and hard to understand. \\ \cmidrule{2-3} 
& 2 & Slightly verbose, unclear logic, needs further simplification. \\ \cmidrule{2-3} 
& 3 & Basically fluent, clear logic, but still room for improvement. \\ \cmidrule{2-3} 
& 4 & Fluent and natural expression, rigorous logic, easy to understand. \\ \cmidrule{2-3} 
& 5 & Very fluent expression, clear and rigorous logic, convincing. \\ 
\midrule

\multirow{5}{*}{Relevance} & 1 & Content is irrelevant to the topic. \\ \cmidrule{2-3} 
& 2 & Content has weak relevance to the topic, needs improvement. \\ \cmidrule{2-3} 
& 3 & Content is relevant to the topic, but some parts need better alignment. \\ \cmidrule{2-3} 
& 4 & Content is highly relevant, covers the topic well. \\ \cmidrule{2-3} 
& 5 & Content is highly relevant, comprehensive and appropriately in-depth. \\ 
\midrule

\multirow{5}{*}{Logic} & 1 & Lacks logic, content is disjointed and hard to follow. \\ \cmidrule{2-3} 
& 2 & Logical relationships are confused, lacks coherence between parts. \\ \cmidrule{2-3} 
& 3 & Basically clear logic, but some jumps are not well-connected. \\ \cmidrule{2-3} 
& 4 & Rigorous logic, good coherence between parts. \\ \cmidrule{2-3} 
& 5 & Very rigorous logic, clear levels, reasonable and strong reasoning. \\ 
\midrule

\multirow{5}{*}{Reference Value} & 1 & Lacks effective reference value, content is trivial or impractical. \\ \cmidrule{2-3} 
& 2 & Low reference value, some parts are unreliable. \\ \cmidrule{2-3} 
& 3 & Certain reference value, but needs further verification. \\ \cmidrule{2-3} 
& 4 & High reference value, accurate and reliable information. \\ \cmidrule{2-3} 
& 5 & Very high reference value, rich and highly credible information. \\ 
\midrule

\multirow{5}{*}{Depth} & 1 & Superficial content, lacks depth, does not address key points. \\ \cmidrule{2-3} 
& 2 & Slightly shallow content, needs deeper exploration of the theme. \\ \cmidrule{2-3} 
& 3 & Certain depth, but some areas need more expansion. \\ \cmidrule{2-3} 
& 4 & Moderate depth, covers most key aspects of the theme. \\ \cmidrule{2-3} 
& 5 & Deep content, comprehensively covers all aspects of the theme. \\ 

\bottomrule
\end{tabular}
\caption{LLM-Based Evaluation Metrics.}
\label{table:Evaluation_Metrics}
\end{table*}

\end{document}